\documentclass[11pt,letterpaper]{article}
\usepackage{naaclhlt2016}
\usepackage{times}
\usepackage{latexsym}

\naaclfinalcopy 

\usepackage{adjustbox}
\usepackage{tikz}
\usetikzlibrary{arrows,positioning} 
\tikzset{
    >=stealth',
    pil/.style={
           <-,
           shorten <=0pt,
           shorten >=0pt,}
}

\makeatletter
\newcommand{\@BIBLABEL}{\@emptybiblabel}
\newcommand{\@emptybiblabel}[1]{}
\makeatother
\usepackage{hyperref}

\usepackage[tone]{tipa}

\usepackage[utf8]{inputenc}
\usepackage[T2A,LAE,T1]{fontenc}
\usepackage[arabic,farsi,USenglish]{babel}

\newcommand{\pword} [1] {\FR{#1}}

\AtBeginDocument{
  \renewcommand\textipa[2][r]{{\fontfamily{cm#1}\tipaencoding #2}}
}

\usepackage{multirow}

\usepackage{mdframed}

\newif\ifcr
\crfalse

\title{MIZ\={A}N: A Large Persian-English Parallel Corpus}

\author{
    Omid Kashefi\\
   Intelligent Systems Program\\
   University of Pittsburgh\\
   {\tt kashefi@cs.pitt.edu}
}

\date{}

\begin{document}

\maketitle

\begin{abstract}
One of the most major and essential tasks in natural language processing is machine translation that is now highly dependent upon multilingual parallel corpora. Through this paper, we introduce the biggest Persian-English parallel corpus with more than one million sentence pairs collected from masterpieces of literature. We also present acquisition process and statistics of the corpus, and experiment a base-line statistical machine translation system using the corpus.
\end{abstract}

\section{Introduction}
Advent of the digital computers in early 20th century revolutionized ways to encounter every aspects of sciences. New interdisciplinary areas, such as corpus linguistic and computational linguistic are destined the automatic translation's state of the art, now referred to as \emph{statistical machine translation} (SMT), that is based on using somehow language independent statistical methods trained by large parallel corpora containing foreign and target language sentence pairs \cite{brown1993mathematics,koehn2003statistical}. 



There exist some multilingual parallel corpora for resource-rich languages such as Europarl \cite{koehn2005europarl} and JRC-Acquis \cite{steinberger2006jrc}. In addition, there are many bilingual corpora, with English as one end in most cases, such as corpora presented in \cite{altenberg2000english,tadic2000building,germann2001aligned,ma2006corpus,utiyama2007japanese}.

However, many limited-resource languages including Persian, lack applicable parallel corpora to benefit from the SMT. First attempt to Persian-English automatic translation was \emph{Shiraz Project} wherein they prepared a parallel corpus with 3K sentence pairs \cite{amtrup2000persian}. Mousavi \shortcite{mosavi2009constructing} proposed a proprietary corpus containing 100K sentence pairs. \emph{TEP} is a publicly available corpus containing about 550K sentence pairs with 8M terms from movies subtitles \cite{pilevar2011tep}. This corpus is built from colloquial Persian that in some cases differs from formal Persian in terms of both morphology and syntax.

Apparently, researchers have attempted to build Persian-English parallel corpora but due to lack of resources and huge amount of required works, the resulted corpora are unsatisfactory in size or quality. Therefore, in order to contribute to Persian-English machine translation research, we present MIZ\={A}N, a manually aligned Persian-English parallel corpus containing 1 million sentence pairs with 25 million terms that is available from \texttt{\small https://github.com/omidkashefi/Mizan}.

We evaluate MIZ\={A}N through a translation task and study how good current SMT approaches are for Persian-English statistical translation and what are the challenges and possible solutions.\\

\section{Corpus Collection}
Parallel contents required for building parallel corpora are usually collected form publicly available texts, mainly from web. However, despite our broad search for available Persian-English parallel texts, we were unable to find enough suitable resources to build our corpus. 

Therefore, searching for any available English text that might have Persian equivalent in any extent, we decide to use copyright-free masterpieces of literature published through \emph{Project Gutenberg} \cite{hart1971project}. We collect a list of 500 titles and look them up in \emph{National Library and Archive of Iran} to see if they have ever been translated into Persian and published in Iran. Among them, about 180 titles were translated to Persian but we find out that most of them were published more that 30 years ago, a fortunate incident, as their copyrights are expired but also a challenge, since they are not available off the shelves. It made us to pursue a cumbersome process of finding used copies one by one.

In parallel to acquiring enough books, we start to digitize them. We decided to use OCR, as a cheap and fast process for digitizing books. However, after working on first 10 titles, we observed that the rate of errors ($WER \approx 30\%$) and the times and expense needed to correct them is such high that it makes the more expensive and slower process of typewriting books reasonable. Therefore, the English side of our comparable text resource was downloaded from Project Gutenberg and the Persian side was manually typewritten from the corresponding translations. Transcription process takes about 3 years employing multiple typists.

\subsection{Refinement}
Refinement is a common preprocessing for SMT \cite{habash2006arabic}. Persian texts suffers vast amount of computational issues from choosing correct character set and encoding to morphological and orthographical ambiguities \cite{kashefi2010towards,rasooliorthographic}.

Persian along with Arabic and Urdu share most of their characters in Unicode. However, there are handful of language dependent but yet homograph exceptions that might mistakenly be used interchangeably. For example, the letter \emph{Yeh} is encoded at U+064A with isolated form representation of \pword{ي}, at U+06CC with isolated form representation of \pword{ی}, and at least encoded in five more places. Using these characters interchangeably forms strings that are computationally different but visually similar that can seriously mislead every statistical analyses. 


Persian language includes three main diacritic classes, \emph{Harekat} that represents short vowel marks (i.e. \pword{ــَــِــُـ}), \emph{Tashdid} that is used to indicate gemination (i.e. \pword{ــّـ}), and \emph{Tanvin} that is used to indicate nunation (i.e. \pword{ــًــٍــٌـ}). The use of diacritics in Persian is not mandatory, however, using diacritics in a word makes it computationally different from that word without diacritics \cite{kashefi2013novel}.


Persian possess intra-word space in addition to inter-word space (i.e. regular white space). An example of intra-word space or pseudo-space is \pword{شرکت‌ها} \textipa{/Serk\ae{}thA:/}, compare to inter-word space as \pword{شرکت ها} and without space as \pword{شرکتها}, all meaning "companies", while two later ones are more common but the former one is correct \cite{kashefi2010optimizing,kashefi2013novel}. 


Challenging these issues we used \emph{Virastyar}\footnote{Virastyar is a free and open-source project, providing fundamental Persian text processing tools. See {\tt http://sourceforge.net/projects/virastyar}}, to correct and normalize non-standard characters based on ISIRI 6219\footnote{\tt{http://www.isiri.org/portal/files/std/6219.htm}}, remove all optional diacritics, unify the \emph{ezafe} usage as short \emph{Yeh} and correct spacing of inflected words.

\subsection{Alignment}
In order to align corresponding sentences of refined books, we developed an alignment aiding software operated by alignment specialists, whom were mostly translators and linguists, to ease the process by providing basic operations such as break, merge, delete and edit tools.

We automatically align corresponding books at chapter level using correspondence score presented in Rasooli \shortcite{rasooli2011extracting}. Then, we change the granularity of alignment to paragraphs and recalculate the score to indicates that the paragraph pairs correspond one-to-one, one-to-two, or not at all. Providing such information warns alignment specialists how much attention and manual work (i.e. break, merge or delete) each paragraph pairs need to ensure alignment. Changing granularity from paragraph to sentence and repeating the same process, we align each parallel books at sentence level.




\subsection{Corpus Statistics}
MIZ\={A}N corpus, containing 1,021,596 unique Persian-English sentence pairs is released in two files encoded in Unicode. Each file contain sentences in a language, each line of files represent a sentence and sentences correspond each other by line numbers. Table \ref{tab:one} shows the number of sentences and words of the corpus on each side.

\bgroup
\def\arraystretch{1.1}
\begin{table}
  \centering
  \small
    \begin{tabular}{lccc}
    \hline\hline
    \textbf{Language}&\textbf{Sentences}&\textbf{Words (Distinct)}\\
    \hline
    Persian&1,011,085&12,049,952 (198,860)\\
    English&1,011,085&11,667,272 (153,666)\\
    \hline
    \textbf{Overall}&\textbf{1,011,085}&\textbf{23,717,224} (\textbf{352,526})\\
    \hline\hline
    \end{tabular}
  \caption{\label{tab:one}Size and statistics of MIZ\={A}N Corpus}
\end{table}
 

\section{SMT Experiment}
To evaluate MIZ\={A}N in a translation task and compare it with currently existing resources, we use Moses toolkit \cite{koehn2007moses}, the available state of the art implementation of phrase-based SMT. We use KenLM \cite{heafield2011kenlm} to build Persian and English language model with order of five. For Persian language model, in addition to Persian side of MIZ\={A}N corpus, we used Hamshahri corpus \cite{aleahmad2009hamshahri} which is a monolingual resource with 10M terms.  


We evaluate the SMT performance using 1,000 held-out and 900 out of domain sentences from an \emph{English in Travel for Persians} (EiT) book. We also build SMT baseline for TEP corpus, which is the only available and the largest Persian-English corpus next to MIZ\={A}N. In order to have a fair comparison, we only compare the result for EiT test set which is out of domain of both MIZ\={A}N and TEP. We tuned each SMT systems with 5,000 in-domain sentences based on minimum error rate.

Table \ref{tab:two} shows the evaluation results of SMT systems in BLEU score. As expected, the base-line SMT model trained by MIZ\={A}N shows superior translation results in terms of BLEU score compared to TEP. Bigger size of MIZ\={A}N corpus along with its higher quality text (formal Persian) and precise manually aligned sentences, caused the differences in translation quality where TEP is mostly colloquial Persian with relatively high number of misaligned sentences that are results of automatic alignment of a highly corresponding comparable text (i.e. movie's subtitles).


\bgroup
\def\arraystretch{1.1}
\setlength\tabcolsep{4.8pt}
\begin{table}
  \centering
  \small
    \begin{tabular}{lccccc}
    \hline\hline
    \multirow{2}{*}&\multicolumn{2}{c}{\textbf{En$\rightarrow$Pr}}&&\multicolumn{2}{c}{\textbf{Pr$\rightarrow$En}}\\
    \cline{2-3}\cline{5-6}
    & Held-out & EiT && Held-out & EiT\\
    \hline
    \textbf{TEP} & - & 6.26 && - & 9.67\\
    \textbf{MIZ\={A}N} & 25.52 & 24.26 && 21.05 & 21.13\\
    \textbf{$\scriptstyle +$Verb} & 27.44 & 25.08 && 22.04 & 21.89\\
    \textbf{Lem} & 26.05 & 25.69 && - & -\\
    \textbf{Lem$\scriptstyle +$Verb} & 27.86 & 26.78 && - & -\\
    \hline\hline
    \end{tabular}
  \caption{\label{tab:two}SMT system performance}
\end{table}

Multiple sentences with different building blocks could express similar concepts, so a sentence could have numerous correct translations. Lets consider the English and Persian sentence pairs shown in Table \ref{tab:four}, along with the translation output made by our SMT system and the post-edited translation result. The BLEU score between translation output and the reference sentence is 26.08, however, with just transposing two first words of the output, the post-edited result is a completely fluent and adequate translation of the English sentence.

\bgroup
\def\arraystretch{1.1}
\setlength\tabcolsep{.1pt}
\begin{table}
    \small
    \begin{tabular}{lr}
       \hline
        \textbf{En:} &\hfill "They listen to their teacher every day"\\
        \textbf{Pr:} &\hfill \pword{آن‌ها هر روز به حرف معلمشان گوش می‌کنند}\\
        \textbf{Output:} &\hfill \pword{به آن‌ها حرف استادشان هر روز گوش می‌دهند}\\
        \textbf{Post-Edited:}&\hfill\pword{آن‌ها به حرف استادشان هر روز گوش می‌دهند}\\
    
       \hline
   \end{tabular}
    \caption{\label{tab:four}An example of BLEU score defect}
\end{table}

Therefore, to get a more clear view about the performance of our SMT system, we asked an expert to post-edit 100 randomly selected output sentences of each test set such each output fluently and adequately satisfy human expectations of a correct translation. We compute BLEU score between translation outputs and their corresponding post-edited translations. Table \ref{tab:five} depict the BLEU score of translation outputs compared with corresponding references and corresponding post-edited results. As it is shown, SMT performance comparing with reference sentences from each test sets varies by 43\%, however, the performance comparing with post-edited results are more consistent and varies by 11\% for different test sets. Moreover, we could roughly infer that our SMT system could satisfies about 50\% of human expectations.

\bgroup
\def\arraystretch{1.1}
\setlength\tabcolsep{8.6pt}
\begin{table}
  \centering
  \small
    \begin{tabular}{lcc}
    \hline\hline
    \multirow{2}{*}{\textbf{Test Set}}&\multicolumn{2}{c}{\textbf{En$\rightarrow$Pr}}\\
    \cline{2-3}
    &{\small References}&{\small Post-Edited Outputs}\\
    \hline
    \textbf{Held-out} & 27.84 & 51.59\\
    \textbf{EiT} & 39.86 & 57.35\\
    \hline\hline
    \end{tabular}
  \caption{\label{tab:five}SMT performance in terms of post-editing}
\end{table}

\subsection{Possible Improvements}
We noticed that majority of inarticulate translations are due to absence or mistranslation of verbs. Investigating extracted aligned phrases, we found out that Moses failed to properly align most of Persian verbs with English verbs. 

Persian is a \emph{subject-object-verb} language while English is a \emph{subject-verb-object}, it means that verbs in English and Persian are differently ordered in sentences. Thus, the long distance relation between corresponding verbs might be the reason of this alignment and translation failure. 

We collect a set of about 300 common English verbs from web and generate their different conjugations. We translate these verbs into Persian and add them to MIZ\={A}N. The result of using combined corpus is reflected in Table \ref{tab:two} on \emph{+Verb} row. Although this is a simple approach but observing higher BLEU score compared to baseline, indicates that considerations for long distance relations and different reordering models need to be studied.

We believe supporting morphology could improve Persian-English SMT performance. We evaluate the effectiveness of morphology-aware SMT by using a simple experiment. We calculate the BLEU score between lemmatized translation outputs and references to avoid the BLUE score drop caused by mismatches of wrong inflections. To this reason we used the morphological analyser and lemmatizer introduced in Virastyar project, that are publicly available. Results are shown in Table \ref{tab:two} on \emph{Lem} row. 

Table \ref{tab:three} shows morphological statistics of test sets. The smaller reduction in output compared to reference, which means outputs have less inflected words than references, shows that our SMT system failed to properly inflect output words. It also implies that supporting morphology will improve SMT performance.

\bgroup
\def\arraystretch{1.1}
\setlength\tabcolsep{4.1pt}
\begin{table}
  \centering
  \small
    \begin{tabular}{lccccc}
    \hline\hline
    \multirow{2}{*} & \multicolumn{2}{c}{\textbf{Reference}} && \multicolumn{2}{c}{\textbf{Output}}\\
    \cline{2-3}\cline{5-6}
    & Held-out & EiT && Held-out & EiT\\
    \hline
    \textbf{Word} & 23,083 & 2,307 && 19,468 & 2,195 \\
    \textbf{Lemmatized} & 4,128 & 549 && 3,451 & 319 \\
    \textbf{Reduction} & 27\% & 23\% && 24\% & 18\% \\
    \hline\hline
    \end{tabular}
  \caption{\label{tab:three}Morphological statistics of datasets}
\end{table}


\section{Conclusion}
In this paper we present MIZ\={A}N, a publicly available and the largest Persian-English parallel corpus with about 1 million sentence pairs. As it has relatively bigger size, more precisely aligned and includes higher quality text compared to other existing and accessible Persian-English corpora, we believe it could be a influential resource for Persian SMT and many other bilingual and cross-lingual corpus related research.

We evaluate the MIZ\={A}N corpus in a SMT task with baseline BLUE score of about 25 for in domain and 24 for simple out of domain test sets. We also investigate some required further studies to improve SMT support for Persian-English and maybe other languages with different word orders.

Nevertheless, Persian language is an under-resource language with many open spots to dig in. Likewise, Persian-English SMT have wide range of open issues and still needs many more research to blossom. 





\section*{Acknowledgments}
This work was supported by a grant from Supreme Council of Information and Communication Technology (SCICT) to the School of Computer Engineering at the Iran University of Science and Technology. 

We also want to thank Dr. Rebecca Hwa, Dr. Mohammad Hedayati Goudarzi, Dr. Behrooz Minaei, Dr. Morteza Analoui, Saaed Alipour, and Ghasem Kasaeian and all of our colleagues who helped us in this project.

\bibliographystyle{naaclhlt2016}
\bibliography{manuscript}

\end{document}